%% file: 0_main.tex
\documentclass[fleqn,12pt]{wlscirep}
\usepackage[utf8]{inputenc}
\usepackage[T1]{fontenc}
\usepackage{hyperref}

\usepackage{breakurl}

\usepackage{amsmath}
\usepackage{graphicx}
\usepackage{epstopdf} 
\usepackage{multirow}
\usepackage{times}
\usepackage{latexsym}
\usepackage{amssymb}
\usepackage{subfigure}
\usepackage{graphicx}
\usepackage[T1]{fontenc}
\usepackage{hyperref} 
\usepackage{bm}
\usepackage[utf8]{inputenc}
\usepackage{hyperref}
\usepackage{booktabs} 

\usepackage{microtype}
\usepackage[lined,boxed,commentsnumbered]{algorithm2e}
\usepackage{inconsolata}

\title{BiomedRAG: A Retrieval augmented Large Language Model for Biomedicine}

\author[1]{Mingchen Li}
\author[2]{Halil Kilicoglu}
\author[3]{Hua Xu}
\author[1]{Rui Zhang}
\affil[1]{Division of Computational Health Sciences, Department of Surgery, University of Minnesota,
Minneapolis, MN, USA}
\affil[2]{School of Information Sciences, University of Illinois at Urbana-Champaign, Champaign, USA}
\affil[3]{Section of Biomedical Informatics and Data Science, School of Medicine, Yale University, New
Haven, CT, USA}
\affil[*]{Corresponding author: Rui Zhang, Ph.D}
\affil[*]{Email: zhan1386@umn.edu}

\input{1_abs}
\begin{document}
\maketitle

\input{2_introduction}

\input{3_results}

\input{4_discussion}
\input{5_method}

\input{7_conclution}

\bibliography{anthology}
\appendix
\section{Appendix}

\end{document}

%% file: 1_abs.tex
\begin{abstract}

Large Language Models (LLMs) have swiftly emerged as vital resources for different applications in the biomedical and healthcare domains; however, these models encounter issues such as generating inaccurate information or hallucinations.  Retrieval-augmented generation provided a solution for these models to update knowledge and enhance their performance.
In contrast to previous retrieval-augmented LMs, which utilize specialized cross-attention mechanisms to help LLM  encode retrieved text, BiomedRAG adopts a simpler approach by directly inputting the retrieved chunk-based documents into the LLM. This straightforward design is easily applicable to existing retrieval and language models, effectively bypassing noise information in retrieved documents, particularly in noise-intensive tasks.
Moreover, we demonstrate the potential for utilizing the LLM  to supervise the retrieval model in the biomedical domain, enabling it to retrieve the document that assists the LM in improving its predictions.  
Specifically, \textsc{BiomedRAG} retrieves relevant documents from a specifically curated, diverse chunk database through a unique, purpose-built chunk scoring mechanism by a tunable scorer. This retrieval process then integrates the selected information directly into the LLM's input, facilitating the generation of expected output, such as structured knowledge. 
Our experiments reveal that with the tuned scorer,\textsc{ BiomedRAG} attains superior performance across 4  biomedical NLP tasks, encompassing information extraction (triple extraction, relation extraction), text classification, and link prediction leveraging over 8 datasets.
For instance, in the triple extraction task,  \textsc{BiomedRAG} outperforms other triple extraction systems with micro-F1 scores of 81.42 and 88.83 on GIT and ChemProt corpora, respectively. 
The outstanding performance underscores the potential of \textsc{BiomedRAG} in constructing effective biomedical intervention systems for various tasks.
\end{abstract}

%% file: 2_introduction.tex
\section{Introduction}
As research in the field deepens, the volume of biomedical literature has grown exponentially. For instance, PubMed is a biomedical literature database, encompassing over 33 million literatures~\cite{hong2020novel}. The widespread adoption of biomedical literature has led to the development and adoption of numerous data mining and statistical techniques. Knowledge extraction technology endeavors to extract structural information from unstructured text~\cite{luo2020neural,lu2022unified}, while link prediction~\cite{li2024condensed,li2022hierarchical} tasks seek to discern the relationship types between two entities, aiding in drug discovery.
To improve support for medical professionals and enhance the performance of BioNLP systems, Biomedical Large Language Models (LLMs)~\cite{ling2023domain} provide a path through pre-training or fine-tuning open-source LLMs in the general domain using biomedical data. This demonstrates notable performance across a range of biomedical tasks. For instance, MedLLaMA~\cite{wu2023pmc} utilized biomedical literature for training and evaluation via biomedical question-answering tasks. GatorTronGPT~\cite{peng2023study} was trained on clinical texts and fine-tuned for various NLP tasks, including biomedical relation extraction, biomedical question answering, etc.

These models are commonly trained on extensive datasets, containing a significant amount of world or domain knowledge implicitly stored within their parameters. However, they are also prone to hallucination~\cite{ji2023survey,zhang2023siren}.
Retrieval-augmented language models~\cite{khandelwal2019generalization,li2023far,lewis2020retrieval,li2023understand}, in contrast, can retrieve knowledge from an external datastore when needed, potentially reducing hallucination and increasing knowledge coverage. 
Previous methods involving retrieval-augmented language models typically require a fixed retriever, such as K-nearest neighbors (KNN)~\cite{taunk2019brief}, to retrieve the most relevant document for the input sentence.  However, these methods mainly perform knowledge retrieval on unlabelled sentences, meaning that the model cannot be guided to learn the (input, label) information. Moreover, in some  noise-intensive tasks, input sentences are likely to introduce words irrelevant to the labels, introducing noise that can adversely affect the performance of the model. For example, in the sentence-level triple extraction task,  consider the sentence \textit{Chitin synthetase \underline{was activated by} fungal acid proteases; however, it was subsequently destroyed by proteases from both animal and plant sources}, the \textit{STIMULATES} relationship between head entity \textit{proteases} and tail entity \textit{synthetase} is discerned from the key chunk \textit{was activated by}. 
The prior retrieval-augmented language models need access to the internal LM representations (e.g., for model training~\cite{borgeaud2022improving}), which poses challenges for their application to very large LMs.
Moreover, numerous state-of-the-art LLMs are only accessible through APIs, with their internal representations undisclosed and lacking support for fine-tuning.

Therefore, to solve the above challenges, in this work,  we primarily investigate the effectiveness of integrating chunk knowledge into LLMs in the biomedical domain, and we propose a novel retrieval-augmented language framework for the biomedical domain, namely \textsc{BiomedRAG},  which is enhanced by a new tailored chunk retriever. 
The key idea is to adapt the retriever to the LLM, which is in contrast to prior work~\cite{shi2023replug} that adapts language models to the retriever. We employ a training objective that prioritizes retrieving chunk-based documents to enhance language model perplexity.
\textsc{BiomedRAG} consists of three major steps: 
(1) constructing the diverse chunk database. In our work, the "chunk" is a broad concept, For instance, in noise-intensive tasks, such as sentence-level relation extraction\footnote{Relation extraction: by giving the sentence and two entities in this sentence, the model needs to extract the relation between these two entities. Triple extraction: by giving the sentence, the model needs to joint extract the triple (head entity, relation, tail entity).} and text classification tasks, the relation type or label typically pertains to several consecutive words in the sentence. Hence, the sentence is divided into multiple chunks. Conversely, in tasks like link prediction, where the input sentence contains only two entities and the output is the relation type, the chunk comprises these two entities.
(2) training the tailored chunk scorer to select the relevant document from the diverse chunk database for the input sentence. 
(3) incorporating the retrieved document into the LLM  to generate the output (e.g. label, structure knowledge, etc.) for the given sentence. 
We perform extensive experiments and demonstrate the effectiveness of our proposed \textsc{BiomedRAG} framework over five tasks ( triple extraction,  relation extraction, text classification, link prediction), showing significant improvement over strong baseline models. The contributions of this work can be summarized as follows:

\begin{itemize}
     \item We proposed \textsc{BiomedRAG}, a new framework that  automatically retrieval chunk information from pre-constructed diverse chunk database for biomedical NLP tasks;
     \item To improve the retrieval quality,  we proposed a learnable tailored chunk scorer to adapt LLM,  utilizing the LLM scores as a supervision signal.
     \item  To assess the model's generalizability, we validated it on 4 biomedical NLP tasks with 8 datasets.
    \item  We conducted a thorough analysis of our method,  including an ablation study,  demonstrating the robustness of our framework.
\end{itemize}

%% file: 3_results.tex
\section{Results}
\label{experiments}
In this section, we present our main results on \textsc{biomedRAG} with a focus on several practical facets, including  1) Comparative evaluations of the \textsc{biomedRAG} framework against other models across 5 tasks and 9 datasets. 2) Comparative assessments with the current RAG model. 
3) Module (tailored chunk scorer, diversity operation) assessment.  4) Model performance under different chunk sizes.

\subsection{Comparative Assessments between Our \textsc{biomedRAG} Framework with Other  Models }
\label{con:discussuion_triple}

Table~\ref{con:Model_performance} and Table~\ref{con:Model_performance_2} present the experiment results on triple extraction, relation extraction, text classification, link prediction. Through our experiments, we noted that our \textsc{biomedRAG} model exhibited the capacity to enhance the performance of  different LLMs such as GPT-4, LLaMA2 13B, and MedLLaMA 13B.

\begin{table*}[ht]
	\centering
	\renewcommand\arraystretch{1.3}
\resizebox{1\textwidth}{!}{%
	\begin{tabular} {l|ccc|ccc|ccc|cc|ccc}
		\toprule 
         \multicolumn{10}{c}  {\textbf{(1) Triple Extraction (TE)}}&\multicolumn{5}{c}  {\textbf{(2) Relation Extraction (RE)}} \\
         \cmidrule(lr){0-9} \cmidrule(lr){11-15}
       
		\multicolumn{1}{c}  {}&\multicolumn{3}{c}  {DDI}& \multicolumn{3}{c}  {ChemProt} & \multicolumn{3}{c}  {GIT} & \multicolumn{2}{c}{}& \multicolumn{3}{c}  {GIT-RE} \\
		 TE Approach & Precision &  Recall & F1 &  Precision &  Recall & F1&  Precision &  Recall & F1& \multicolumn{2}{l}{RE Approach}& Precision &  Recall & F1 \\
	      \midrule
		UniRel~\cite{tang2022unirel}  & 25.59  & 21.92& 27.13    &29.00 & 17.45&21.79 &36.36  &  12.50  & 18.60 &\multicolumn{2}{l}{RT-10~\cite{li2023far} }& 42.80   &43.44   &   43.12 \\
        OneRel~\cite{shang2022onerel} &34.72  & 81.07  & 22.09    &44.95  &45.22  &  44.68 & 47.97 &  52.01  &44.52& \multicolumn{2}{l}  {RT-5~\cite{li2023far}} &44.91  & 46.45 &  45.67\\
                UIE (base)~\cite{lu2022unified} & 30.74  &  29.07 & 29.88   &  48.15 &44.79  &46.41  & 34.81 &30.59    &32.56 &\multicolumn{2}{l}  {RT-1~\cite{li2023far} }& 44.86  &  46.02 & 45.44 \\ 
                 E2H (large)~\cite{gao2023easy}  & 30.24  &  30.16  &    30.20  &  49.86   & 47.78  & 48.80  &31.47  &  26.89  & 29.00  & \multicolumn{2}{l}  { RT-20~\cite{li2023far} } &44.94   & 45.81  & 45.37 \\
                E2H (base)~\cite{gao2023easy} &  34.23 &   34.25 & 34.24   & 48.92   & 46.51 &47.68    & 31.29 &  26.68  &28.80 &  \multicolumn{2}{l}  {BERT~\cite{devlin2018bert}} & 86.25& 86.25 & 86.25 \\
               UIE (large)~\cite{lu2022unified} &  37.50  & 35.05& 36.24   &    49.74& 46.27    &  47.94 & 33.99 &  29.93  & 31.83  & \multicolumn{2}{l}  {BioBERT~\cite{lee2020biobert} }&85.43 & 85.43 &85.43 \\
                 GPT-4  &  5.08    &    9.00&  6.50    &  20.79   &37.00   &     26.61& 9.48 & 9.46   &9.47  & \multicolumn{2}{l}  {GPT-4} &48.17 &48.17 & 48.17\\

                      MedLLaMA 13B~\cite{wu2023pmc} &  76.63    &  76.63      &  76.63      & 52.10   &49.04   & 50.52   &  42.60& 41.51  & 42.05   &\multicolumn{2}{l}  {MedLLaMA 13B~\cite{wu2023pmc} }& 67.02   & 66.88 & 66.95 \\
                   LLaMA2 13B~\cite{touvron2023llama} &    79.61    &79.61     & 79.61       &   78.58  & 76.41 & 77.48    &61.76   &56.45    &58.99  &\multicolumn{2}{l}  { LLaMA2 13B~\cite{touvron2023llama}}&79.43 & 78.92  & 79.18\\
                    
                    \hline
                 MedLLaMA 13B + \textsc{biomedRAG} &   76.90    &76.90    &76.90    &77.48   & 76.46& 76.97  & 76.88 & 76.56   &  76.72 &\multicolumn{2}{l}  {MedLLaMA 13B + \textsc{biomedRAG}} &86.66&86.66&86.66\\
                 LLaMA2 13B + \textsc{biomedRAG} &  \textbf{80.50} &   \textbf{79.10 } &     \textbf{80.00} &   \textbf{89.25}  &  \textbf{88.42} &  \textbf{88.83}  & \textbf{81.78} & \textbf{81.07}   &\textbf{81.42} & \multicolumn{2}{l}  {LLaMA2 13B + \textsc{biomedRAG}} & \textbf{89.03}&\textbf{89.03}&\textbf{89.03} \\
           \bottomrule
	\end{tabular}
 }
\vspace{+2mm}
\caption{Results of various approaches on triple extraction, relation extraction over 4 datasets. }
\label{con:Model_performance}
\vspace{-1mm}
\end{table*}

\begin{table*}[ht]
	\centering
	\renewcommand\arraystretch{1.3}
\resizebox{1\textwidth}{!}{%
	\begin{tabular} {l|ccc|ccc|ccc|ccc}
		\toprule 
                  \multicolumn{7}{c}  {\textbf{(1) Text Classification}}& \multicolumn{6}{c} {\textbf{(2) Link Prediction}}\\ 
        \cmidrule(lr){0-6} \cmidrule(lr){8-13} 
        \multicolumn{1}{c}  {}&\multicolumn{3}{c} {Ade-corpus-v2}& \multicolumn{3}{c}{MTsample} &\multicolumn{3}{c}{UMLS}& \multicolumn{3}{c}{ADInt}\\
		 TE Approach & Precision &  Recall & F1 &  Precision &  Recall & F1&  Precision &  Recall & F1& Precision &  Recall & F1\\
        \hline
           BERT~\cite{devlin2018bert}&   95.00   & 97.00  &  96.00    &  38.00 &  38.00  &    38.00  & 57.00   &  57.00  &  57.00  & 58.15 &58.15  &  58.15 \\
        BioBERT~\cite{lee2020biobert}&  95.00   & 97.00   &   97.00   &  37.00& 37.00 & 37.00& 58.00   &  58.00  & 58.00  & 59.37&  59.37 &   59.37\\
            GatorTron~\cite{yang2022gatortron}&   95.00    &  98.00  &  97.00    &  38.00 & 38.00 & 38.00    & 59.00   &  59.00  &   59.00  &59.23 &59.23  & 59.23 \\
                \hline
               MedLLaMA 13B~\cite{wu2023pmc} &  95.40 &  95.40  &  95.40   &24.27 & 23.14  &  23.69   & 34.51   &32.37   &  33.41  &46.25   & 46.25   & 46.25    \\
              LLaMA2 13B~\cite{touvron2023llama} &  96.40  &   96.40  &   96.40    &  38.49  & 36.80  &  37.62   & 56.12    & 56.12     &  56.12  &  61.38 & 61.38 &61.38   \\

             GPT-4&   41.00   &    41.00   &   41.00     &   41.33 & 41.33  & 41.33 &  6.00   &    6.00  &  6.00     & 33.33  & 33.33  &33.33 \\
            \hline
            MedLLaMA 13B + \textsc{biomedRAG} &  \textbf{99.89}  &  \textbf{99.89} &   \textbf{99.89}  &   32.52  &  32.45 & 32.49  & 58.00  &  58.00  &  58.00 &59.16 &59.16 & 59.16 \\
             LLaMA2 13B + \textsc{biomedRAG} & 99.80 &  99.80   &  99.80    &  38.50  &38.50 &38.50  &\textbf{59.80} &    \textbf{59.80} & \textbf{59.80}  &\textbf{62.22}& \textbf{62.22} & \textbf{62.22}\\
             GPT-4 + \textsc{biomedRAG}&   75.60   &    75.60   & 75.60       &     \textbf{41.93} &   \textbf{41.93 }&   \textbf{41.93}  &   24.35   &   24.35   &   24.35       &37.08    &  37.08  & 37.08   \\
           \bottomrule
	\end{tabular}
 }
\vspace{+2mm}
\caption{Results of various approaches on  text classification, link prediction over 4 datasets. }
\label{con:Model_performance_2}
\vspace{-1mm}
\end{table*}

\begin{table*}[ht]
	\centering
	\renewcommand\arraystretch{1.3}
\resizebox{1\textwidth}{!}{%
	\begin{tabular} {l|l|ccc|ccc|ccc|ccc}
		\toprule 
  
		\multicolumn{1}{c}  {}&\multicolumn{1}{c}  {}&\multicolumn{3}{c}  {Triple}& \multicolumn{3}{c}  {Head Entity}& \multicolumn{3}{c}  {Tail Entity} & \multicolumn{3}{c}  {Relation}  \\
		 &Approach & Precision &  Recall & F1 &  Precision &  Recall & F1 &  Precision &  Recall & F1 &  Precision &  Recall & F1 \\ 

	      \midrule
                
             \multirow{2}*{DDI}&UIE (large)&37.50   & 35.05  & 36.24 &  68.89  &64.40   & 66.57 &  63.37  & 59.24 & 61.23 & 45.63   &  42.66 & 44.10   \\
               &LLaMA2 13B + \textsc{biomedRAG}& \textbf{80.50} &  \textbf{79.10} & \textbf{80.00 }& \textbf{97.01}  &\textbf{97.01}  & \textbf{97.01} & \textbf{ 85.59 } &   \textbf{85.59}&\textbf{85.59} &\textbf{92.11}   & \textbf{ 92.10 }&   \textbf{92.12} \\
              \hline
             \multirow{2}*{ChemProt}&E2H (large)&49.86  &  47.78 & 48.80 & 80.45  & 77.09 & 78.73 & 78.23    &74.97   & 76.56&  71.26 & 68.29  &  69.74  \\
               &LLaMA2 13B + \textsc{biomedRAG}& \textbf{89.25} & \textbf{88.42} & \textbf{88.83} & \textbf{98.26}  & \textbf{97.35}  &\textbf{97.80 }  & \textbf{96.71}   &\textbf{95.81}   & \textbf{96.25} &  \textbf{93.01} & \textbf{92.14}  &    \textbf{92.57}\\
                \hline
             \multirow{2}*{GIT}&OneRel &47.97  & 52.01  &  44.52  & 65.26 & 71.54 &60.00   &   68.16   & 75.46  &62.15 & 74.70 & 82.94 & 67.96  \\
            &LLaMA2 13B + \textsc{biomedRAG}&  \textbf{81.78}   & \textbf{81.07 }& \textbf{81.42} & \textbf{92.84} &  \textbf{92.04}  & \textbf{92.44}    &\textbf{91.76}    &\textbf{90.96} & \textbf{91.36}   &   \textbf{87.20} & \textbf{86.45} & \textbf{86.82}   \\
             
           \bottomrule

	\end{tabular}
 }
\vspace{+2mm}
\caption{Triple, head entity, tail entity and relation results of various approaches on DDI, ChemProt and GIT}
\label{con:Model_entity_relation_triple}
\vspace{-5mm}
\end{table*}

\paragraph{Triple Extraction}
In our paper, triple extraction is defined as the task where the model extracts the triple (head entity, relation, tail entity) from sentences.
From Table~\ref{con:Model_performance} (1), we have the following observations: On GIT, (1) our \textsc{biomedRAG} significantly outperforms all the strong baselines across all evaluation metrics. (2) we observe that  \textsc{biomedRAG} improve the original    MedLLaMA 13B and LLaMA2 13B  by  34.67\%, 22.43\%  respectively, in terms of Triple-F1. 
(3) The performance of lightweight models like UniRel and OneRel significantly lags behind that of the LLaMA2 family, like LLaMA2 7B. 
On DDI and ChemProt,
we have the following observations: (1) our \textsc{biomedRAG} still outperforms all baselines across F1 value on these open datasets.  (2) We observe that  \textsc{biomedRAG } improve the original    MedLLaMA 13B and LLaMA2 13B  by 0.27\%, 0.39\%  respectively  on the DDI.  (3) We observe that  \textsc{biomedRAG } improve the original    MedLLaMA 13B and LLaMA2 13B  by 26.45\%, 11.35\%  respectively  on the ChemProt. 
(4) The lower performance observed in Unirel and Onerel can be attributed to the table-filling method struggling to recognize complex biomedical entities or relations. For example, as illustrated in Table~\ref{con:Model_entity_relation_triple}, on DDI, UIE achieves only a 61.23\% accuracy for tail entity recognition. 

Table~\ref{con:Model_entity_relation_triple} presents the entities,  relation types, and triple comparisons among the top-1 baseline models from Table~\ref{con:Model_performance} and \textsc{biomedRAG}. We observed that  (1) \textsc{biomedRAG} achieved state-of-the-art performance in triple F1, relation F1 and entity F1. (2) For the lightweight model, such as UIE, it achieves a lower F1 score for relations on the DDI dataset. (3) 
Lightweight models like UIE and OneRel struggle with entity recognition in biomedical sentences.
To further validate the performance of our model, we present the results of triple extraction from several state-of-the-art systems: SPBERE~\cite{yang2023spbere} and JBUIM~\cite{tan2023joint}, as sourced from their respective papers. Since their code is not publicly available, we compare their model performance solely on the DDI and ChemProt datasets. Our findings indicate that our model continues to demonstrate superior performance compared to these systems.
On the DDI dataset, JBUIM and SPBERE achieved F1 scores of 77.70\% and 79.20\% respectively, whereas our model achieved an F1 score of 80.00\%. Similarly, on the ChemProt dataset, JBUIM and SPBERE achieved F1 scores of 68.80\% and 69.70\%, while our model achieved an F1 score of 88.83\%.



\paragraph{Relation Extraction} In our paper, the relation extraction task involves predicting the type of relationship between two entities (head entity and tail entity) based on a given sentence, head entity, and tail entity. To assess the scalability of our model, in this part, we evaluate the performance of \textsc{biomedRAG} in relation extraction tasks. 
Table~\ref{con:Model_performance} (2) presents the experiment results of various approaches. We have the following observations: (1) our \textsc{biomedRAG} significantly outperforms all the strong baselines and its variants across all evaluation metrics. (2) We observed that \textsc{biomedRAG } improve the original   \textsc{MedLLaMA 13B }, and \textsc{LLaMA2 13B }   by  19.71\%, and 9.85\%  respectively, in term of F1.  
(3) Without training, GPT-4 struggles to extract the relationship in the sentence.

\paragraph{Text Classification}
Table~\ref{con:Model_performance_2} (1) presents the experiment results of various approaches on the two datasets ade-corpus-v2 and MT-sample.   (1) It's notable that \textsc{biomedRAG} enhances the original \textsc{MedLLaMA 13B}, \textsc{LLaMA2 13B} and GPT-4 by  4.49\%, 3.40\% and 34.60 respectively, in terms of F1 score.  (2) Without fine-tuning, GPT-4 exhibits lower performance on the ade-corpus-v2 dataset. Nevertheless, \textsc{biomedRAG} has the capability to enhance the performance of GPT-4, surpassing the performance achieved by BERT, BioBERT, and GatorTron.
On MTSample, we observed that: (1) GPT-4, without fine-tuning, achieves the best performance when compared to models like LLaMA2 13B and MedLLAMA 13B, which require fine-tuning. (2) Remarkably, \textsc{biomedRAG} boosts the performance of the original \textsc{MedLLaMA 13B}, \textsc{LLaMA2 13B}, and \textsc{GPT-4} models by 8.80\%, 0.88\%, and 0.60\%, respectively, in F1 score.

\paragraph{Link Prediction} Table~\ref{con:Model_performance_2} (2)  presents the experiment results of various approaches on the UMLS and ADInt.  On UMLS, we can see that: (1)  \textsc{biomedRAG} improve the original  \textsc{MedLLaMA 13B },  \textsc{LLaMA2 13B } and \textsc{GPT-4}   by 24.59\%,  3.68\%  and  18.35\% respectively, in term of F1. 
 On ADInt,  (1) we observed that \textsc{biomedRAG} improve the original  \textsc{MedLLaMA 13B}, \textsc{LLaMA2 13B }   and \textsc{GPT-4} by 12.91\%,  0.84\% and 3.75\%  respectively, in term of F1. 
 

\subsection{Comparative Assessments with RAG models}
\label{con:topn-sentence_retrival}
\begin{figure*}[htbp]
    \centering
    \subfigure[DDI]{
        \includegraphics[width=2in]{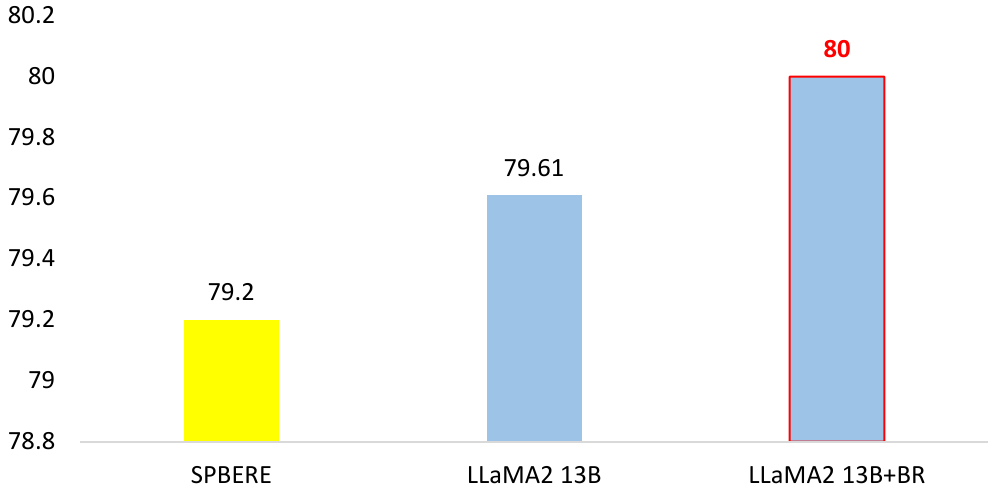}
    }
    \subfigure[ChemProt]{
	\includegraphics[width=2in]{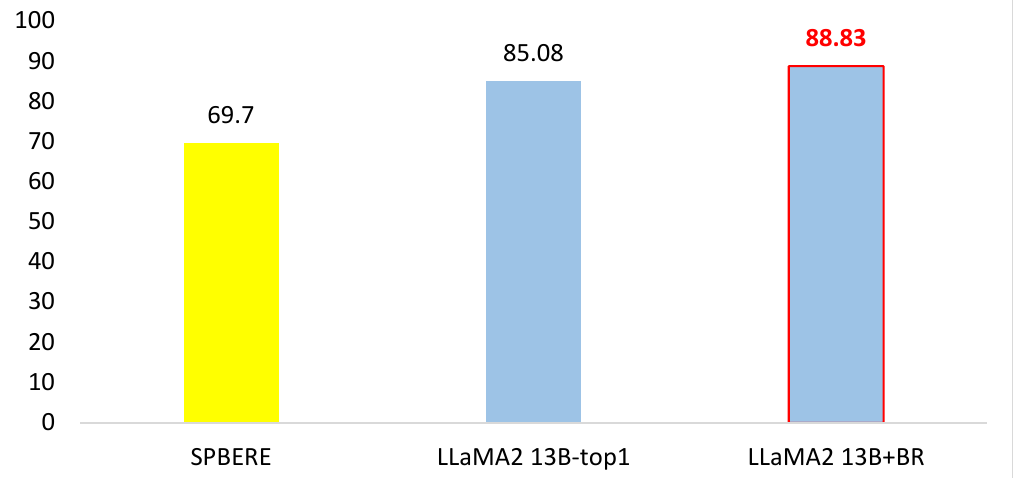}
    }
  \subfigure[GIT]{
	\includegraphics[width=2in]{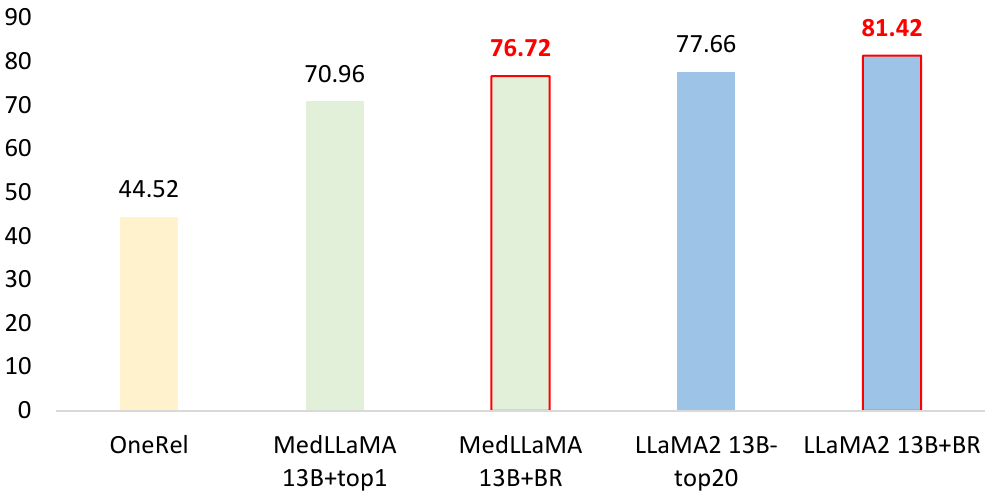}
    }
 \subfigure[GIT-RE]{
	\includegraphics[width=2in]{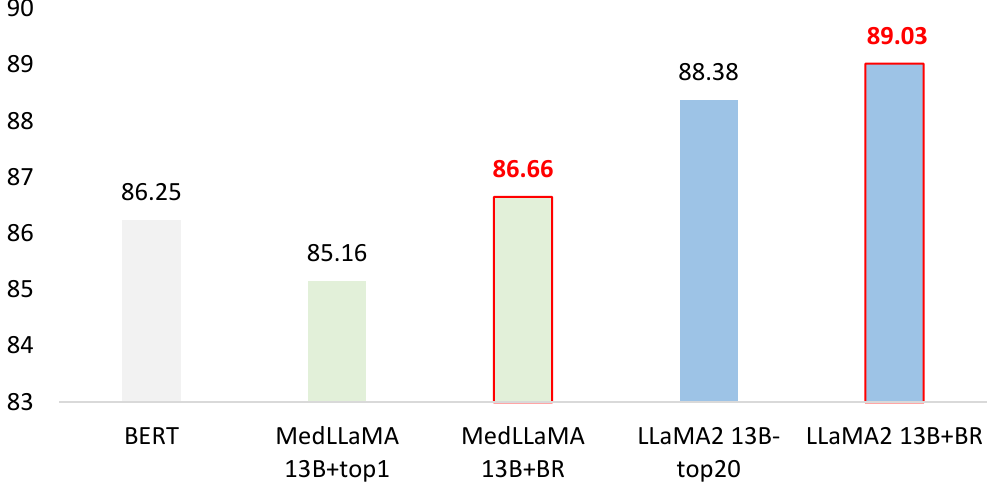}
    }
 \subfigure[Ade-corpus-v2]{
	\includegraphics[width=2in]{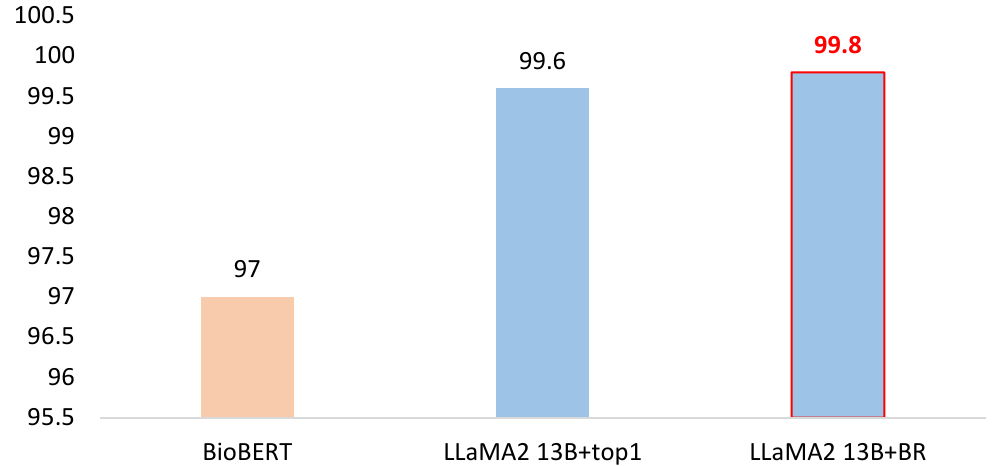}
    }
 \subfigure[MTsample]{
	\includegraphics[width=2in]{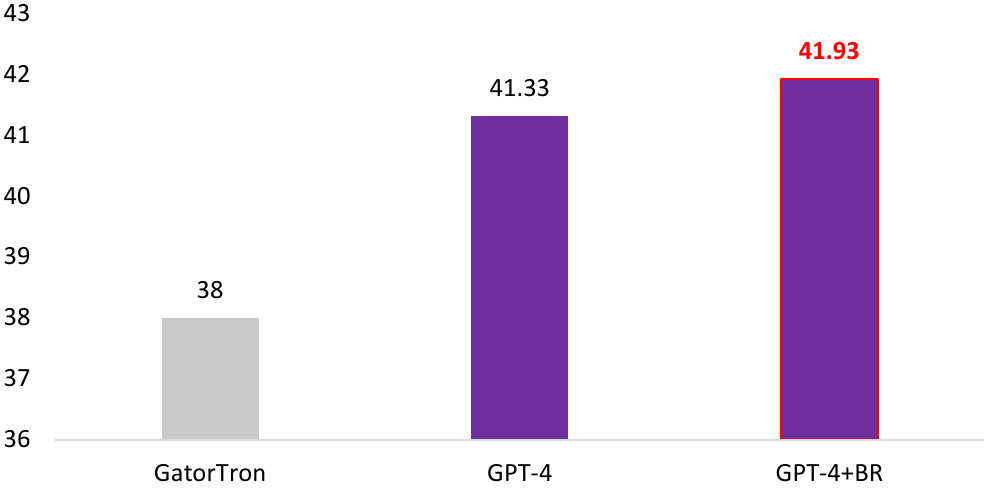}
    }
     \subfigure[UMLS]{
	\includegraphics[width=2in]{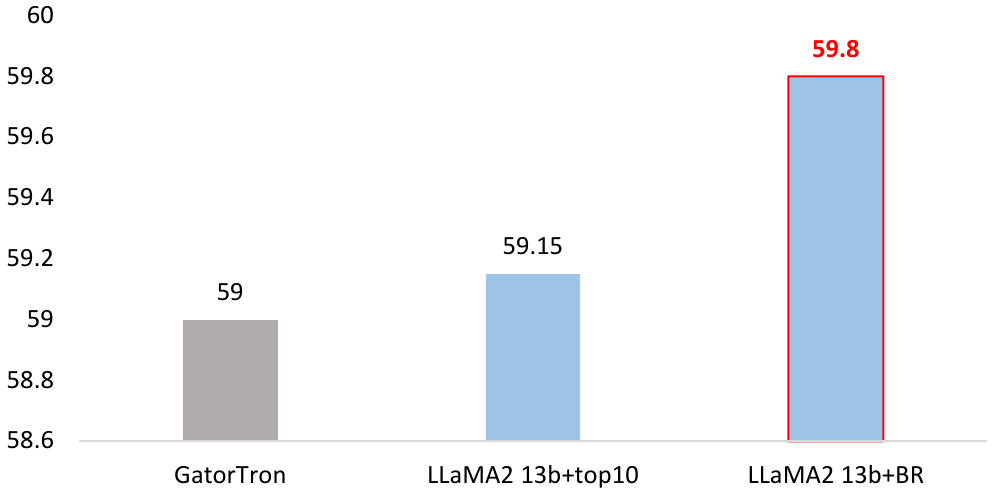}
    }
   \subfigure[ADInt]{
	\includegraphics[width=2in]{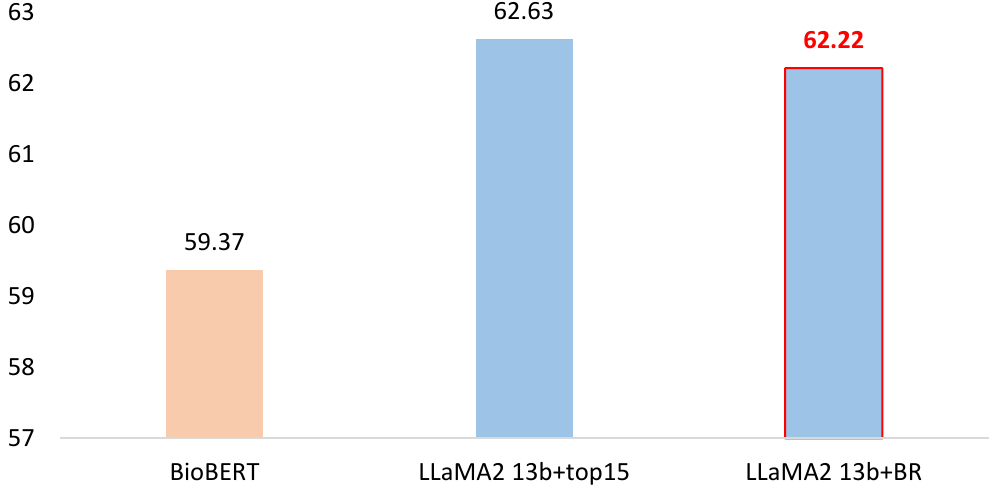}
    }
    \caption{F1(a-h)/Accuracy(i) performance of different models. BR refers to \textsc{biomedRAG}. The red font indicates the performance of \textsc{biomedRAG}.}
    \label{con:best_baseline_model_results}
\end{figure*}

The core of our method design lies in establishing a relational key-value memory at the chunk level and training a tailored chunk scorer to adapt the LLM. To comprehensively assess our model's performance, we conduct a comparative analysis with the prevailing retrieval-based LLM, employing a retriever to obtain the top-$n$ relevant documents related to the input sentence. we called this model as RA-\textsc{KNN}-$n$,The results from the top-performing baseline models and the \textsc{bimedRAG} are illustrated in Figure \ref{con:best_baseline_model_results}. 

\begin{figure*}[htbp]
    \centering
    \subfigure[DDI (LLaMA2 13B)]{
        \includegraphics[width=2in]{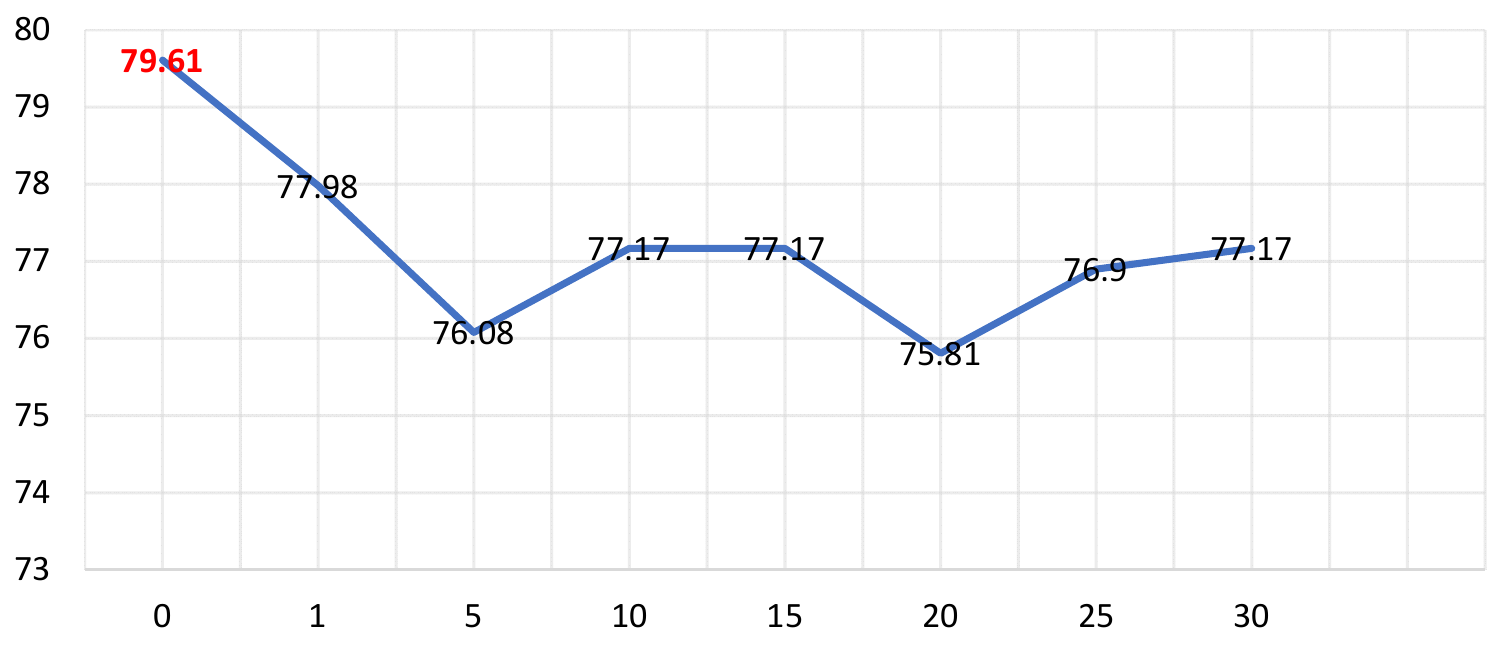}
    }
    \subfigure[ChemProt (LLaMA2 13B)]{
	\includegraphics[width=2in]{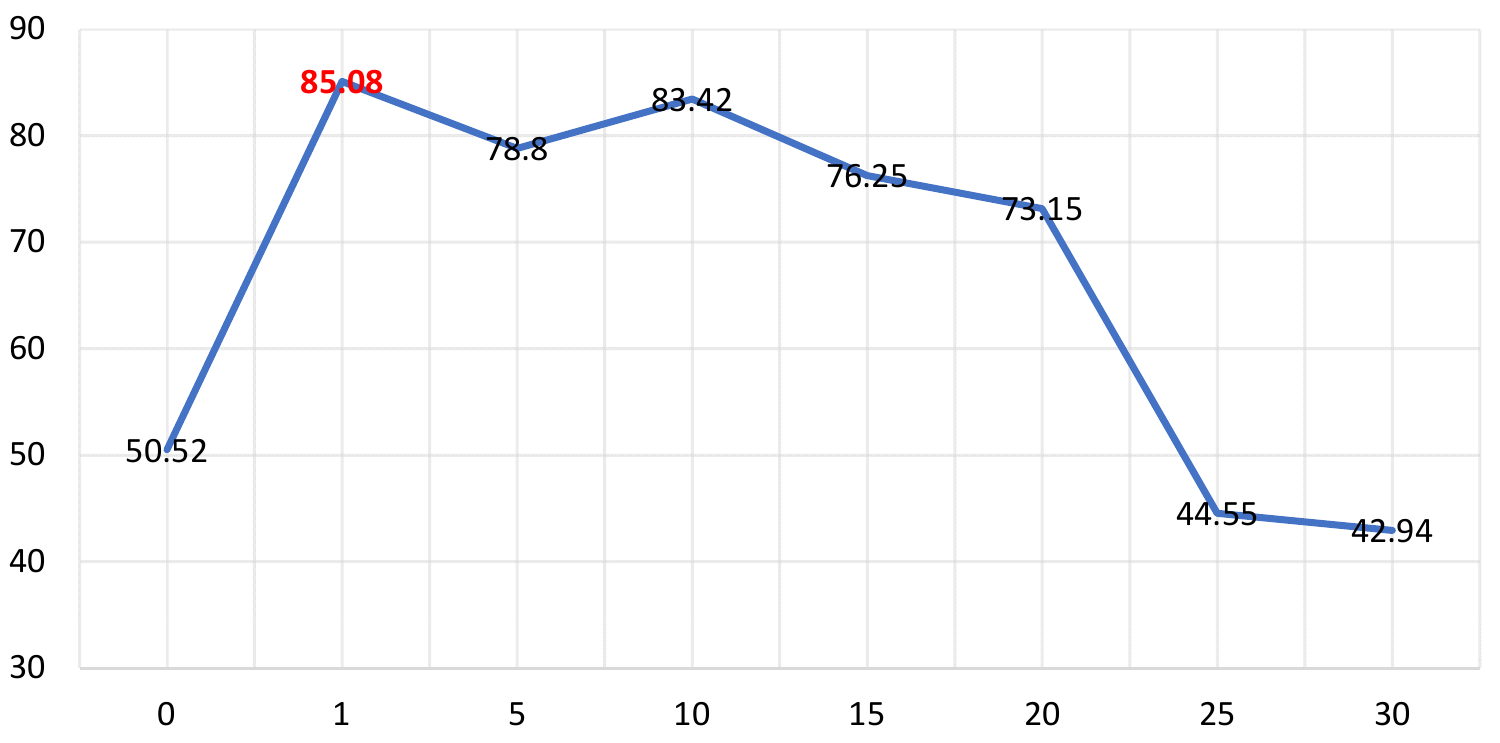}
    }
    \subfigure[GIT \& GIT-RE (MedLLaMA 13B)]{
	\includegraphics[width=2in]{ 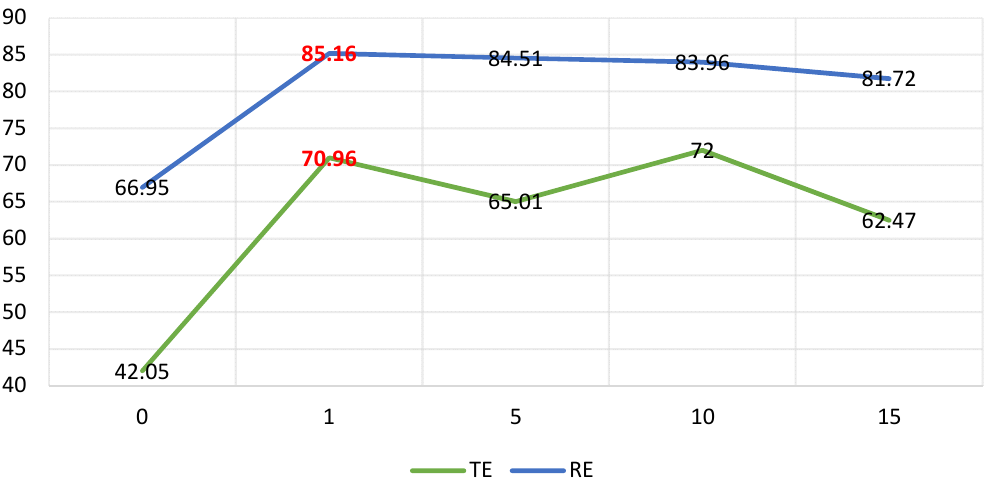}
    }
 \subfigure[GIT\& GIT-RE (LLaMA2 13B)]{
	\includegraphics[width=2in]{ 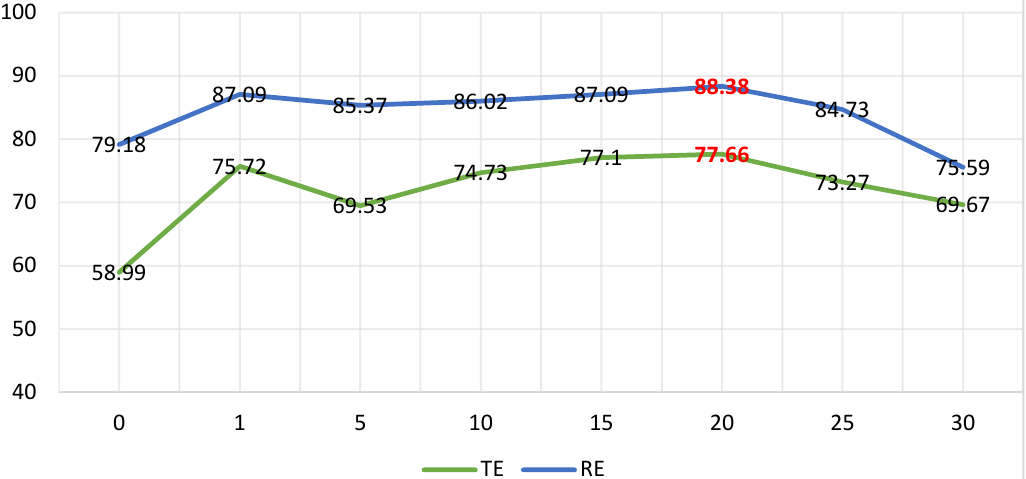}
    }
     \subfigure[Ade-corpus-v2 (LLaMA2 13B)]{
	\includegraphics[width=2in]{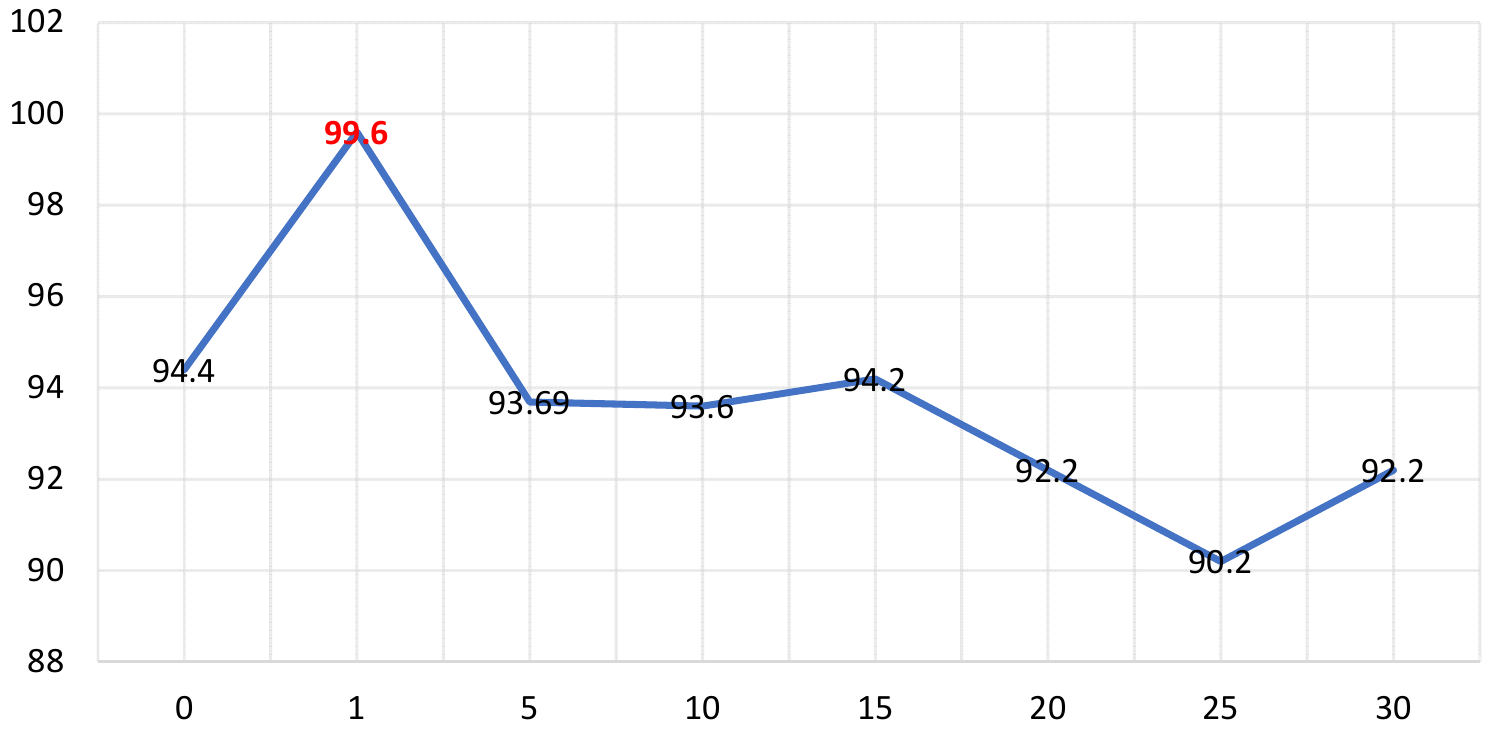}
    }
 \subfigure[MTsample (GPT-4)]{
	\includegraphics[width=2in]{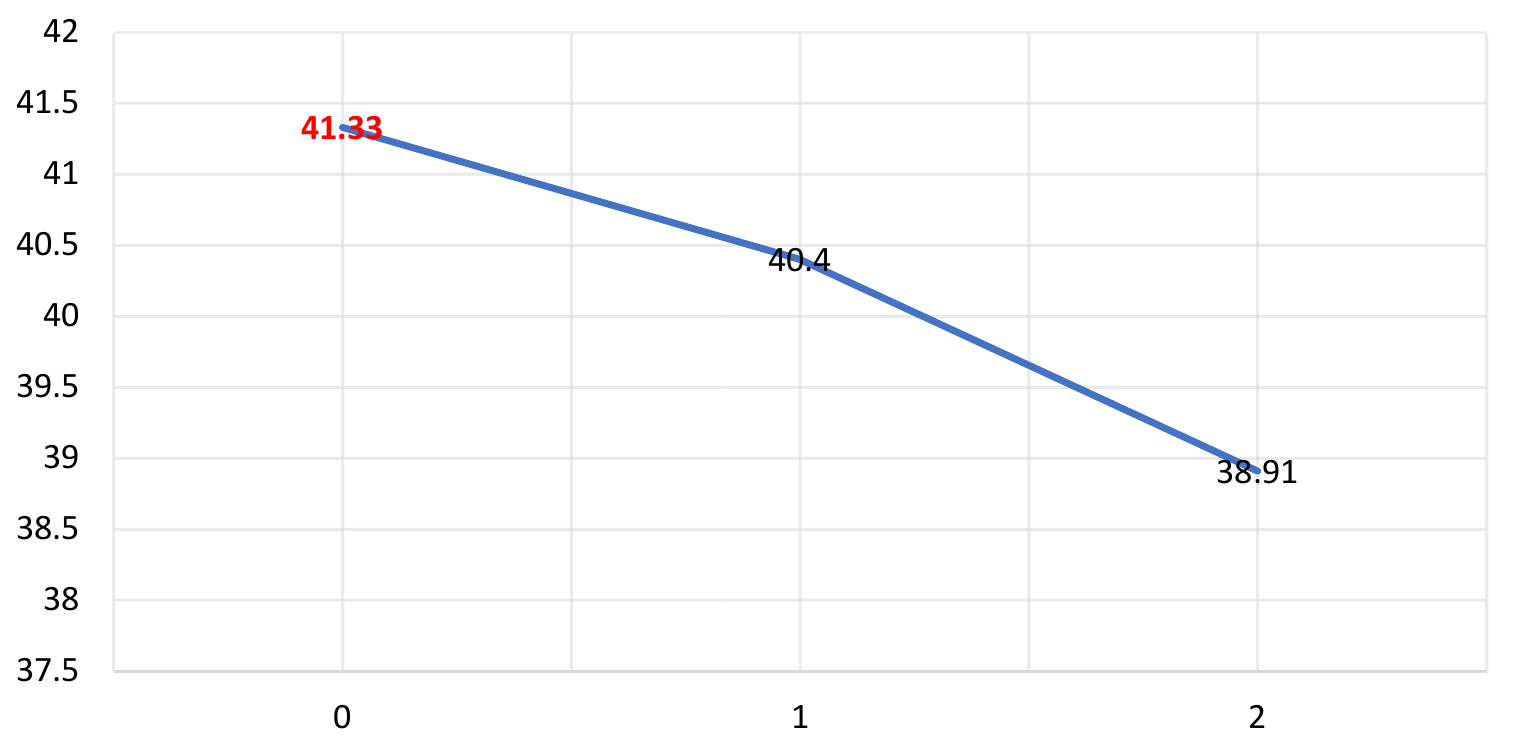}
    }
 \subfigure[UMLS (LLaMA2 13B)]{
	\includegraphics[width=2in]{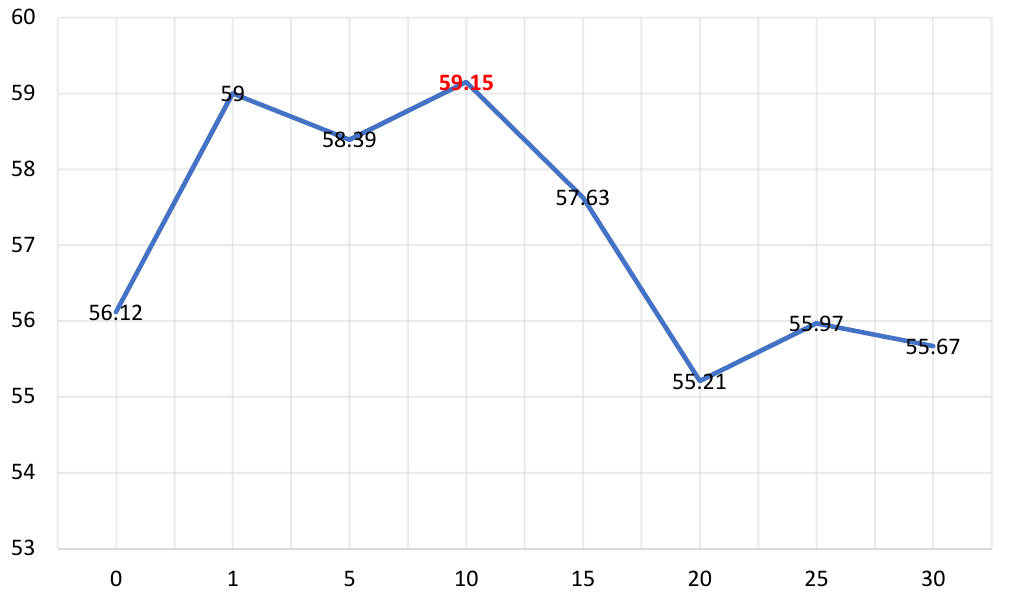}
    }
    \subfigure[ADInt (LLaMA2 13B)]{
	\includegraphics[width=2in]{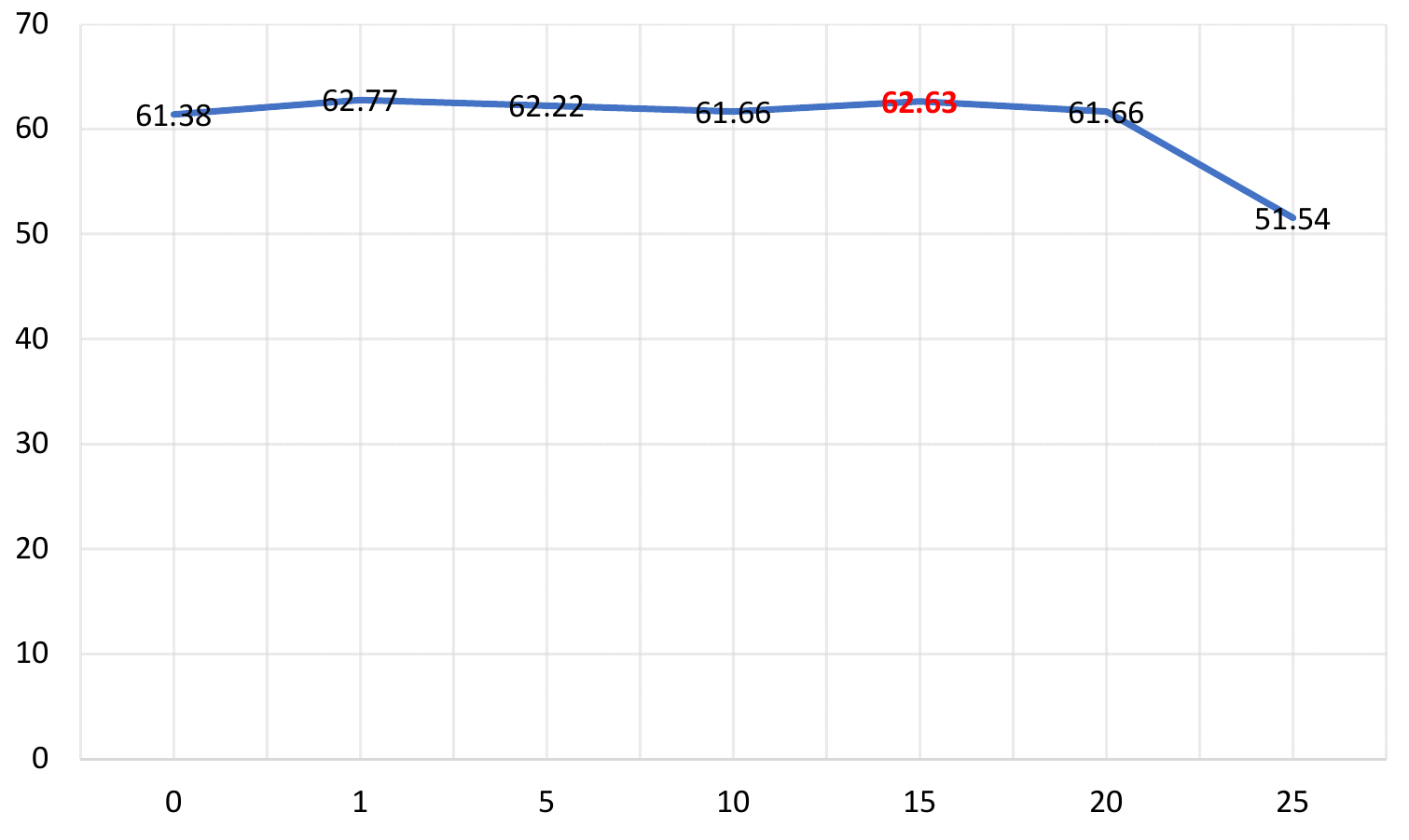}
    }

    \quad    
    \caption{Different F1 (a-h)/Accuracy (i) performance of  KNN-based retrieval model on 5 tasks over 9 datasets. We select the LLM with the highest baseline performance as our primary LLM model. The number of document samples depends on the input length constraint of the LLM. For instance, in dataset Mssample, when $top-n>2$, the maximum input token length of 8318 exceeds the maximum input token length of 8192 tokens for GPT-4.   y-axis: F1 (a-h)/Accuracy (i). x-axis: 0 represents no retriever, while 1-30 represents the top-n documents retrieved. The red font refers to the best performance. }
    \label{con:KNN_retrival}
\end{figure*}

More specifically, same as \cite{li2023far, ram2023context}, we employ K-nearest neighbors (KNN) as the retriever to obtain the top-n (sentence, label) pairs from the retrieval database (same with the retrieval database as our method). The model results of different $n$ are shown in Figure \ref{con:KNN_retrival}.

In the experiment of GIT and GIT-RE, we selected LLaMA2 13B and MedLLaMA 13B as  LLM. Due to input length constraints, we present the performance results for the top-15 in MedLLaMA 13B and top-30 in LLaMA2 13B. As shown in  Figure~\ref{con:best_baseline_model_results}, upon comparing the F1 values between OneRel and MedLLaMA 13B+top1, we observe that the retrieval-based model proves effective in enhancing triple extraction performance.
As depicted in Figure~\ref{con:KNN_retrival}(a) and Figure~\ref{con:KNN_retrival}(b), we observed that the value of $n$ in top-$n$ does not exhibit a direct proportionality with the LLM's performance. Notably, as shown in Figure~\ref{con:best_baseline_model_results}(c-d), with the incorporation of \textsc{biomedRAG}, we observed an enhancement in the performance of both MedLLaMA 13B and LLaMA2 13B beyond the best achieved through retrieval alone. 
Otherwise, we observed that even though our model retrieves only the top-1 document from the diverse chunk database, it still outperforms the KNN-based RAG-LLM in retrieving the top-20 documents, For instance, in Figure \ref{con:best_baseline_model_results} (c), \textsc{bimedRAG} achieves 89.03\%, whereas \textsc{LLaMA2 13B}+top20 only achieves 88.38\% on GIT (retrieving top-20 document from the train set). 
It is also crucial to highlight that while retrieving more sentences can improve the performance of the original LLM, it also leads to increased training time. Despite the superior performance of LLaMA2 13B+top15 compared to LLaMA2 13B+BR, as demonstrated in Figure~\ref{con:best_baseline_model_results}, retrieving larger documents will impact both training and inference times, along with consuming additional computational resources.

From Figure~\ref{con:KNN_retrival}(a,b,c,d,e,f,g,h), We've discovered that retrieving as much as documents is not necessary to enhance the model's performance, for example,  Figure~\ref{con:KNN_retrival}(b), as when $n=30$, the model gets a worse performance. For both the triple extraction (Figure~\ref{con:KNN_retrival}(c)), text classification task  (Figure~\ref{con:KNN_retrival}(e)) and relation extraction tasks (Figure~\ref{con:KNN_retrival}(c)), we observed that the model achieves its best performance when $n=1$. Retrieving documents appears to be beneficial for enhancing label prediction accuracy.
In Figure~\ref{con:KNN_retrival}, we observe that not all tasks benefit from using the LLM with the KNN-based retrieval method. For instance, on the DDI dataset, the RA-KNN-n model exhibits lower performance compared to the LLM without example-guided generation. However, our \textsc{biomedRAG} demonstrates superior performance when compared to both the RA-KNN-n and other base models.

Consequently, our model also proves effective in reducing training time associated with retrieving additional sentences and addressing challenges related to input length limitations.


\subsection{Model Performance under Different Chunk Sizes}

\begin{figure}[htbp]
    \centering
    \subfigure[]{
        \includegraphics[width=3in]{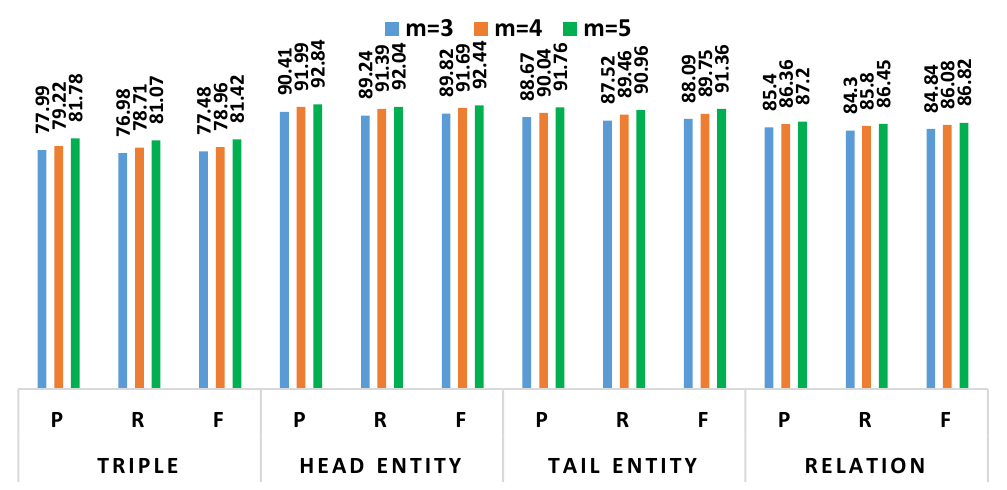}
    }
    \subfigure[]{
	\includegraphics[width=3in]{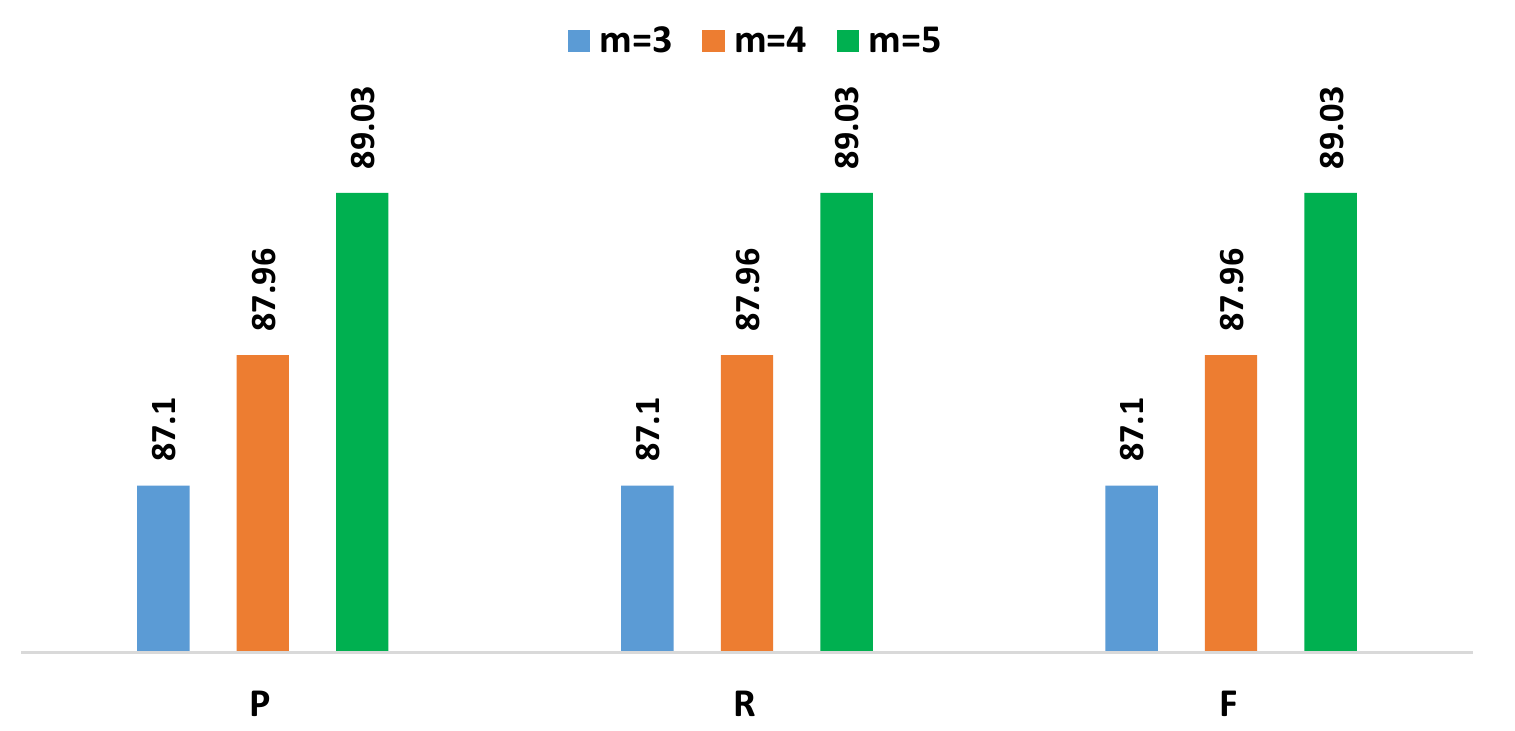}
    }
    \quad    
    \caption{(a): Precision (P), Recall (R), F1 results with different chunk length $m$ settings in the task of biomedical triple extraction task. (b): Precision (P), Recall (R), F1 results with different chunk length $m$ settings in the task of biomedical relation  extraction task. }
    \label{con:chuck_number}
\end{figure}

In our proposed model, there is a parameter that controls the number of chunks: $m\in \{3, 4, 5\}$ when constructing the relational key-value memory, which represents the length of each chunk. This parameter aids in retrieving relevant chuck information, considering that the input sentence often contains noise that can impact generation progress.  In this section, we show the model performance on two tasks triple extraction and relation extraction—that have shown significant improvement by \textsc{biomedRAG}, as shown in Figure~\ref{con:chuck_number}(a) and Figure~\ref{con:chuck_number}(b). 
As we can see, in these two tasks, when $m=5$, \textsc{biomedRAG} can construct the more rich relation-based information, and retrieve the more relevant relation chunk to the input sentence.  In addition, the performance of $m=5$ is better than $m=3$ or $m=4$, also indicating that if the granularity of relation chunk is too coarse, it will impact relation recognition and consequently affect model performance.

\subsection{Module (tailored chunk scorer, diversity operation) assessment}

\begin{table}[ht]
	\centering
	\renewcommand\arraystretch{1.3}
\resizebox{0.5\textwidth}{!}{%
	\begin{tabular} {l|l|ccc}
		\toprule 
		 &Approach & Precision &  Recall & F1 \\
            \hline
        \multirow{3}*{Triple Extraction}& LLM + \textsc{biomedRAG}&  \textbf{81.78}   & \textbf{81.07 }     & \textbf{81.42} \\
        & LLM+\textsc{biomedRAG }-WTCS&  79.31   &79.14       &79.22     \\ 
        &LLM+\textsc{biomedRAG }-WD&  75.93   &  74.62     & 75.27     \\ 
        \hline
        \multirow{3}*{Relation Extraction}&LLM + \textsc{biomedRAG}&    \textbf{89.03}  &   \textbf{89.03 }  &   \textbf{89.03} \\
        &LLM+\textsc{biomedRAG }-WTCS &86.66   &86.66      &  86.66 \\  
        &LLM+\textsc{biomedRAG }-WD & 85.80  &85.80      & 85.80  \\ 
           \bottomrule
	\end{tabular}
 }
\vspace{+2mm}
\caption{\textsc{biomedRAG} performance about its ablated model. we chose the LLaMA2 13B as
the LLM.}
\label{con:ablated models}
\vspace{-5mm}
\end{table}

The contribution of our model components can also
be learned from ablated models.  In this part, we mainly focus on biomedical triple extraction and relation extraction task, which get significant improvement over our \textsc{biomedRAG}. We introduce two
ablated models of \textsc{biomedRAG}, (1) \textsc{biomedRAG}-WTCS
uses a  cosine similarity to choose the most relevant document from diverse chunk database
, without using the Tailored Chunk Scorer.
(2) \textsc{biomedRAG}-WD does not consider the diversity of the chunk database.
In these experiments, we selected LLaMA2 13B + \textsc{biomedRAG} as it demonstrated the highest performance from Table~\ref{con:ablated models}. 
 We find that the performance of \textsc{biomedRAG} degrades as we remove
important model components. Specifically, as shown in Figure.~\ref{con:ablated models} both \textsc{biomedRAG}-WTCS and \textsc{biomedRAG}-WD  perform poorly
when compared to \textsc{biomedRAG}, indicating the importance of training a tailored chunk scorer to adapt the LLM and improve the diversity of chunk database.

%% file: 4_discussion.tex
\section{Discussion}
In this paper, we introduce \textsc{biomedRAG}, a novel approach aimed at enhancing the performance of original Language Learning Models (LLMs) such as MedLLaMA 13B, LLaMA2 13B, and GPT-4. Our proposed method leverages the utilization of retrieved chunk documents, acquired through a specifically trained tailored chunk scorer, to augment the capabilities of LLMs.
This improvement highlights the efficacy of integrating retrieved chunk documents into the LLM framework. For example, by comparing our model with the current RAG model as shown in Figure~\ref{con:KNN_retrival}, we observe that even when retrieving only the top-1 document from the diverse chunk database, our model outperforms the KNN-based RA-LLM in retrieving the top-n documents. This suggests that \textsc{biomedRAG} excels in retrieving diverse chunk documents by utilizing a tailored chunk scorer, thereby significantly enhancing the overall performance of the model.

On the triple extraction task, GPT-4 demonstrates notably lower performance, particularly in the case of GIT. The main reason is that the reported results are in the zero-shot setting due to the unavailability of open resources. 
UniRel performs significantly worse than OneRel, primarily due to the superiority of the BIE operation in OneRel compared to the Interaction Map in UniRel for entity recognition.  However, the lower performance observed in Unirel and Onerel can be attributed to the table-filling method struggling to recognize complex biomedical entities or relations. For instance, Onerel achieves only 62.15\% for tail entity recognition on GIT. It remains challenging to recognize some complex biomedical words.  
In the realm of triple extraction tasks, GPT-4 exhibits markedly diminished performance, particularly evident in its performance on GIT. This decline can be chiefly attributed to the fact that the reported outcomes stem from a zero-shot setting, necessitated by the absence of accessible resources.
UniRel finds itself notably outpaced by OneRel, primarily owing to the superior efficacy of the BIE operation within OneRel in contrast to the Interaction Map employed in UniRel for entity recognition. Nonetheless, the subdued performance witnessed in both UniRel and OneRel can be ascribed to the limitations of the table-filling method in accurately identifying intricate biomedical entities or relationships. For instance, OneRel achieves a meager 62.15\% accuracy for tail entity recognition on GIT, underscoring the persistent challenge of accurately recognizing certain complex biomedical terms.

It is noteworthy that during experimentation on both GIT and GIT-RE, we observed intriguing results when varying the chunk number parameter. Specifically, when setting $m=5$, \textsc{biomedRAG} demonstrated an enhanced capacity to construct richer relation-based information and retrieve more relevant relation chunks corresponding to the input sentence. Moreover, the performance achieved with $m=5$ surpassed that of $m=3$ or $m=4$, highlighting the detrimental impact of overly coarse granularity on relation recognition and subsequently on model performance.
Furthermore, the efficacy of this particular chunk number extends beyond the GIT dataset, proving beneficial in tackling noise-intensive tasks and datasets such as ChemProt, DDI, and Ade-corpus-v2. This underscores the versatility and effectiveness of utilizing an optimal chunk number parameter in enhancing model performance across various domains and challenging datasets.

In both the text classification task and link prediction, we made the intriguing observation that the lightweight language model demonstrated performance on par with the larger language model. This parity may stem from the lightweight model's adeptness at handling general tasks, coupled with the absence of particularly challenging input sentences necessitating intricate semantic parsing.
Notably, our exploration of the MTsample dataset revealed a fascinating phenomenon: even without fine-tuning, GPT-4 exhibited superior performance compared to the larger language model such as LLamA2 13B, even when fine-tuned. We postulate that GPT-4's ability to harness its inherent knowledge for performance enhancement during prompt tuning contributes to this phenomenon.

%% file: 5_method.tex
\section{Method}

\subsection{Datasets}

In this section, we describe the dataset we used in our paper, Table~\ref{con:data_Statistics_DDI_chemprot} shows the data statistics for GIT, CHEMPROT and DDI.   Table~\ref{con:others} shows the data statistics for ade-corpus-v2,  MTsample, UMLS, ADInt.

\subsubsection{Triple Extraction Dataset}  
In this paper, we utilized GIT, DDI, and Chemprot as the foundational datasets. 

\begin{itemize}
  \item GIT~\cite{li2023petailor} is a high-quality biomedical triple extraction dataset for non-drug
therapies, characterized by its high-quality annotations and comprehensive coverage of relation types. It includes 22 relation types from SemMedDB.
  \item CHEMPROT~\cite{taboureau2010chemprot}: The Chemical Protein Interaction Corpus comprises 2432 PubMed abstracts annotated with chemical-protein interactions, encompassing 23 distinct interaction relations. Building upon prior research~\cite{sun2022mrc4bioer}, the corpus exclusively considers sentence-level instances, with a particular focus on five prominent interaction types for classification: CPR3, CPR4, CPR5, CPR6, CPR9.
  \item DDI~\cite{segura2013semeval}: The DDI dataset was formulated to support drug information extraction (IE) research for SemEval 2013. It comprises 233 texts sourced from Medline abstracts and 792 texts from the DrugBank database. This dataset encompasses four distinct types of relations between drug entities, namely Advice, Mechanism, Effect, and Int.
\end{itemize}

\begin{table}[ht]
	\centering
	\renewcommand\arraystretch{1.3}
	\scalebox{0.55}{
	\begin{tabular} {ccccc}
		\hline 
		Dataset& \# Entities &  \#Relation Types&\# train/test/dev \\ 
		\hline	
      CHEMPROT~\cite{taboureau2010chemprot}& 5,990&  5 & 4,111/3,438/2,411   \\
        DDI~\cite{segura2013semeval}&13,107  & 4 &   5,154/1,376/1,436  \\
    GIT\cite{li2023petailor}   & 5,644 &22 &  3,734/465/492 \\
  
		\hline
	\end{tabular}
 }
		\caption{Data Statistics for CHEMPROT, DDI, and GIT. "train/test/dev" denotes the counts of (sentence, triples) pairs within each training, testing, and development dataset split.}
	\label{con:data_Statistics_DDI_chemprot}
\end{table}

\begin{table}[ht]
	\centering
	\renewcommand\arraystretch{1.3}
	\scalebox{0.55}{
	\begin{tabular} {cccc}
		\hline 
		Dataset& train &test& dev\\ 
		\hline	
      ade-corpus-v2~\cite{GURULINGAPPA2012885} & 4,000  &500&500   \\
        MTsample~\footnote{https://mtsamples.com/} & 4,029 &500& 500  \\
      ADInt~\cite{xiao2023repurposing}  &6,000   &720&720  \\
       UMLS~\cite{kok2007statistical}  & 5,216  &661&652 \\
		\hline
	\end{tabular}
 }
		\caption{Data Statistics for  ade-corpus-v2, MTsample, UMLS,ADInt. "train/test/dev" denotes the counts of (sentence, triples) pairs within each training, testing, and development dataset split.}
	\label{con:others}
\end{table}

\subsubsection{Relation Extraction}
In this paper, we utilized GIT-RE as the foundational dataset in relation extraction task. 
The biomedical triple extraction dataset GIT serves as the source data for the relation extraction task. We convert the GIT dataset into the relation extraction dataset GIT-RE by using the input context, head entity, and tail entity as model inputs, where the model's output signifies the relationship between the input head entity and tail entity.

\subsubsection{Text Classification}
In this paper, we utilized ade-corpus-v2~\cite{GURULINGAPPA2012885}   and  MTsample~\footnote{https://mtsamples.com/} as the foundational dataset in the text classification task. 
\begin{itemize}
   \item  ade-corpus-v2 dataset is designed for classifying whether a sentence is ADE( Adverse Drug Reaction)-related (True) or not (False). In our paper, we randomly select 4,000 instances for training, 500 for testing, and 500 for validation.
   \item  The MTsample dataset, aims to understand the nature of the language used in medical transcriptions of various kinds. It includes more than 40 classes.
\end{itemize}

\subsubsection{Link Prediction}
In this paper, we utilized UMLS~\cite{kok2007statistical}    and  ADInt~\cite{xiao2023repurposing} as the foundational dataset in the link prediction task. 
\begin{itemize}
   \item  UMLS~\cite{kok2007statistical} contains triples from the Unified
Medical Language System, providing knowledge in the domain of healthcare
and medicine. It comprises 6,529 triples, divided into 5,216 for training, 652 for validation, and 661 for testing.  
   \item  ADInt~\cite{xiao2023repurposing} is a dataset for identifying new pharmaceutical interventions (PI) for Alzheimer's Disease (AD).   In our paper, we randomly select  6,000 samples from the source training set for training, 720 samples from the source testing set for testing, and 720 samples from the source validation set for validation.
\end{itemize}


\subsection{\textsc{Our method: BiomedRAG}}

\begin{figure*}[t]
        \centering
        \includegraphics[width=1\columnwidth]{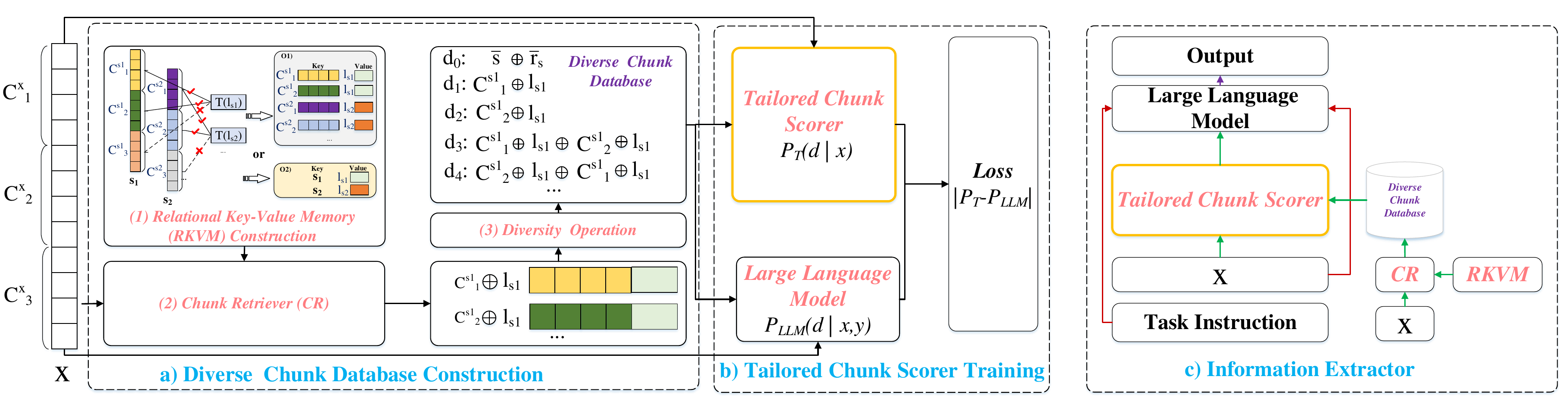}
	\caption{ Overview of \textsc{BiomedRAG}.}
	\label{con:whole_framework}
\end{figure*}

Figure~\ref{con:whole_framework} gives an overview our \textsc{BiomedRAG}, which consists of three major parts: 
\paragraph{(a)  Diverse Chunk Database Construction} involves three substeps:
1) Develop the Relational Key-Value Memory (RKVM) $M$.
2) Employ the Chunk Scorer to retrieve the relevant key-value  (chuck-label) pairs corresponding to the input sentence. 
3) Construct the diverse chunk database through diversity operations, incorporating relevant key-value pairs along with the most pertinent (sentence, label) pair extracted from the validation dataset for each input sentence.

\paragraph{(b)Tailored Chunk Scorer Training} 
The Tailored Chunk Scorer's training process centers on choosing the most relevant document from a diverse chunk database based on a given input sentence, with guidance from the LLM scores.

\paragraph{(c) Information Extractor} 
The Information Extractor generates the output (e.g. relation type, structural knowledge, etc.) by combining the input with the document that has the highest weight score from the diverse chunk database.

\subsubsection{Diverse Chunk Database Construction}

\textbf{RKVM  Construction:}
In this part, we introduce how to construct the key-value memory based on chunks to aid in biomedical applications. For noise-intensive tasks, such as text classification,  sentence-level relation extraction, the chunk refers to the several consecutive words divided into a sentence. Specifically,  we utilize the validation data to aid in constructing the source dataset $D=\{s,l_s\}$, where $s$ denotes the sentence and $l_s$ signifies the label of sentence $s$, for example, if this task is relation extraction, $l_s$ refers the relation type of the entity pair in sentence $s$.

Subsequently, the sentence $s$ is divided into $v$ chunks, each with a length of $m=w/v$, where $w$ represents the length of the $s$. Following this, we compute the similarity between each label $l_s$ and the $v$-th  chunk $C^s_v$ in sentence $s$ using the cosine similarity $S(,)$, as follows:

$$sim(T(l_s),C^s_v)=S(\mathbf{E}(T(l_s)),\mathbf{E}(C^s_v))$$

where $T(l_s)$ is the text description of  label  $l_s$. 
$\mathbf{E}(.)$ is the encoder function, we used MedLLaMA 13B~\cite{wu2023pmc} as $\mathbf{E}(.)$ in our work. 
Subsequently, for each \textit{value} $l_s$ in $M$, we determine its associated \textit{key} by selecting the top two chunks from the sentence $s$.  RKVM $M$  can be defined as:


$$M_r^s=\{ (\underbrace{ C^s_{1}}_{key}, \underbrace{l_s}_{value}),(\underbrace{ C^s_{2}}_{key},\ \underbrace{l_s}_{value})\}$$
$$ M=\{M_r^s\}$$
Where, $C^s_{1}$ and $C^s_{2}$ represent the top-1 and top-2 chunks, respectively, for the label $l_s$ within the sentence $s$.
For example, in Figure~\ref{con:whole_framework}, the sub memory $M_{l1}^{s1}$ pertaining to label $rl$ regarding sentence $s1$ can be defined as follows:  $(C^{s1}_{1}, l_{s_1})$ and $(C^{s1}_{2}, l_{s_1})$.

When constructing $M$ for none noise-intensive tasks, like link prediction, the chunk (key) denotes the input (head entity, tail entity) pair, with the value representing the relation. 
In our work, the noise-intensive task includes triple extraction, relation extraction, and text classification, while none noise-intensive tasks include link prediction.

\textbf{Chunk Retriever:}
The chunk retriever aims to get the most relevant  \textit{key-value} ($k$-$v$)  pairs from  RKVM for the $i$-th split chunk $C^x_i$  which has the same length as $C_v^s$ in input sentence $x$.
More specifically, we use MedLLaMA 13b as our encoder to map each \textit{key} $k$ and chunk $C^x_i$ to the embeddings $\mathbf{E}(k)$ and $\mathbf{E}(C^x_i)$. The similarity between the  chunk embedding and \textit{key} embedding is computed by the cosine similarity:
$$sim(k, C^x_i)=S(\mathbf{E}(k),\mathbf{E}(C^x_i))$$
Subsequently, the \textit{key} with the highest similarity, along with its corresponding \textit{value}, will serve as the retrieved key-value pair $a_i$ for chunk $C^x_i$. So the retrieved  key-value pairs $A_x$ for the input sentence $x$ can be represented by:


$$A_x=\{a_0,a_1,...,a_i\}, a_i=k_i \bigoplus v_i$$
For tasks that aren't heavily reliant on noise, $A_x$ comprises the top-n relevant key values in $M$. In our paper, we set $n=10$. For example, if the task is a classification task, the $A_x$ includes the top $n$ \{key (sentence), value (label)\} pairs.
\subsubsection{Diversity Operation}
Retrieving diverse knowledge has shown the potential to enhance the generation capabilities in NLP tasks, as observed in tasks like Dialogue State Tracking~\cite{king2023diverse}. In our study, we assume that diverse knowledge has the potential to offer more meaningful and rich information to guide the LLM  and enhance its ability in expected output identification. Therefore, we use permutation as the diversity operation in our work.
Specifically, the permutation operation is specifically applied to $A_x$. The resulting permutation of $A_x$, combined with the $(\overline{s}, \overline{l_s})$ pair, constitutes the Diverse Chunk Database $\hat{A_x}$.
Here, $(\overline{s}, \overline{l_s})$ is selected from the source dataset $D=\{({s},{l_s})\}$, demonstrating the highest cosine similarity with $x$.

$$\hat{A_x}=\{\underbrace{ \overline{s} \bigoplus \overline{l_s}}_{d_0}, a_0, a_1,...,\underbrace{..a_{i-1} \bigoplus a_i}_{d_j}\}$$

An example  of permutation operation on $A_x$,  \begin{small}
$$A_x:(C_1^{s1} \bigoplus l_{s1}, C_2^{s1} \bigoplus l_{s1}) \xrightarrow{} \hat{A_x}: (\overline{s}\bigoplus \overline{l_s}, C_1^{s1} \bigoplus l_{s1}, C_2^{s1} \bigoplus l_{s1}, C_1^{s1} \bigoplus l_{s1}\bigoplus  C_2^{s1} \\ \bigoplus l_{s1}, C_2^{s1} \bigoplus l_{s1}\bigoplus C_1^{s1} \bigoplus l_{s1})$$
\end{small}
Note that if the number of chunks is too large, the permutation operation will impact the training time, and the model will be affected by the maximum length limitation of the language model. So in this situation, $\hat{A_x}=\{\underbrace{ \overline{s} \bigoplus \overline{l_s}}_{d_0}, A_x\}$.

\subsubsection{ Tailored Chunk Scorer Training}

\textbf{Tailored Chunk Scorer}:
There exists a different weight value between input sentence $x$ and document $d_j$ in  $\hat{A_x}$.
Tailored Chunk Scorer aims to learn the weight value between input context $x$ and each $d_j$. 
%
Specifically, the input sentence $x$ and each document $d_j$ in $\hat{A_x}$ are encoded into the sentence embedding $\mathbf{E}(x)$ and document embedding $\mathbf{E}(d_j)$. We then calculate the similarity of each document $d_j$ by:


$$P_T(d_j|x)=\frac{e^{sim(x,d_j)/\eta} }{\sum_{d_j \in \hat{A_x}} e^{sim(x,d_j)/\eta}}$$

Where $\eta$ represents a hyperparameter that regulates the temperature of the softmax function. The document retrieval likelihood $P_T(d|x)$
is calculated by calculating the highest probability between document $d_j$ and input sentence  $x$.

\textbf{Training the Tailored Chunk Scorer}:
We use the LLM as a scoring function to help train the  Tailored Chunk Scorer and measure how much each document $d_j$ could improve the LLM perplexity.
In the training process, the $d_j$ that makes the LLM's output as close as possible to the ground truth is considered to be providing the document that the LLM needs. Specifically, we first compute $P_{LLM}(y|d_j,x)$, the LLM probability of the ground truth $y$ given the input sentence $x$ and a document $d_j$. The higher the probability, the better the document $d_j$ is at improving the LLM's perplexity. So we compute the LM likelihood of each document $d_j$ as follows:

$$P_{LLM}(d|x,y)=\max(P_{LLM}(y|d_1,x),...,P_{LLM}(y|d_j,x))$$

The  Tailored Chunk Scorer is trained by minimizing the loss function between  the document retrieval likelihood  and LM likelihood:


$$L=\frac{1}{\mid B \mid }\sum_{x\in B}\mid P_T(d|x)- P_{LLM}(d|x,y)$$

\subsubsection{ Information Extractor}
\label{Information_Extractor}
We construct the instruction-based datasets for each biomedical application. Specifically, the dataset contains four components: 
1) Instruction ($I$), a manually defined guide for the LLM to generate output (such as, triples) for each sentence. 2) Example, where we employ the trained tailored chunk scorer to assign weights to each document in $\hat{A_x}$ for each input context $x$. The document $\bar{d_j}$ with the highest weight score is considered as the example.
3) Input sentence $x$. 4) output $t$.  The expected output $t$ in our model is predicted by the function,
$$P(t|x)=P(t|I \bigoplus \bar{d_j}\bigoplus x) $$
In the generation progress, instruction $I$, example $\bar{d_j}$, and input sentence $x$ are fed into the LLM to generate the  $t$ of the $x$. 

\subsection{Baselines}

To validate the effectiveness of our framework \textsc{BiomedRAG}, this section describes the baselines employed across various tasks.
\subsubsection{Triple Extraction}
We selected six open triple extraction models as the baseline for the triple extraction task, 
they are  UniRel~\cite{tang2022unirel}, OneRel~\cite{shang2022onerel}, UIE (base)~\cite{lu2022unified}, UIE (large)~\cite{lu2022unified},  E2H (base)~\cite{gao2023easy}, and E2H (large)~\cite{gao2023easy}. 
We also assess the performance of \textsc{BiomedRAG} by comparing it with several robust baselines built on state-of-the-art LLMs, including 
 \textbf{1) LLaMA family} as baselines, namely MedLLaMA 13B~\cite{wu2023pmc} and LLaMA2 13b~\cite{touvron2023llama}.  
\textbf{2) GPT-4}, we formulate prompts to guide the GPT-4 models in generating triples for each input sentence, along with providing the corresponding relation definitions in the prompts. \textbf{3) Retrieval-argumented Large Language Model (RA-\textsc{KNN}-$n$)}.  As shown in Figure~\ref{con:KNN_retrival}, we used the MedLLaMA 13B and LLaMA2 13B as the base model. To further assess the efficacy of \textsc{BiomedRAG}, we also employed two state-of-the-art biomedical triple extraction models as baselines on the DDI and ChemProt datasets. Their respective results from their source papers include  JBUIM~\cite{tan2023joint}, and SPBERE~\cite{yang2023spbere}. As the unavailability of the code, we did not report the results on GIT.





\subsubsection{Relation Extraction}
We compare the performance of \textsc{BimedRAG} with several strong baselines based on the LLM, including   \textbf{1) GPT-4}:  In GPT-4, we design prompts to guide the GPT models in predicting relations between the head and tail entities for each input sentence.  \textbf{2) RT-n}:  We also employ RT~\cite{li2023far}, a retrieval-augmented and chain-of-thoughts method, to extract the relation between the head and tail entity. $n$ refers the top-$n$ documents. Same as the baselines in our triple extraction task, we consider \textbf{3) LLAMA family} as the baselines: MedLLaMA 13B~\cite{wu2023pmc}, LLaMA2- 13b~\cite{touvron2023llama}.  
\textbf{4) RA-\textsc{KNN}-$n$}: it is consistent with the   RA-\textsc{KNN}-$n$ in the baseline of the triple extraction task.  As shown in Figure~\ref{con:KNN_retrival}, we used the MedLLaMA 13B and LLaMA2 13B as the base model.
\textbf{5)BERT~\cite{devlin2018bert} and BioBERT~\cite{lee2020biobert}}.
\subsubsection{Text Classification}
On the  text classification task,  we compare the performance of \textsc{BimedRAG} with several strong baselines based on the language models, including   1) \textbf{BERT~\cite{devlin2018bert} and  BioBERT~\cite{lee2020biobert}}, 2) \textbf{GatorTron}~\cite{yang2022gatortron},  we also consider 3) \textbf{LLAMA family} as the baselines: MedLLaMA 13B~\cite{wu2023pmc}, LLaMA2-13B~\cite{touvron2023llama}.   4)\textbf{GPT-4}. 
\textbf{5) RA-\textsc{KNN}-$n$}: As shown in Figure~\ref{con:KNN_retrival}, 
in Ade-corpus-v3, LLaMA2 13B serves as the base model, while in MTsample, GPT-4 serves as the base model.

\subsubsection{Link Prediction}
On the link prediction task,  we compare the performance of \textsc{BimedRAG} with several strong baselines based on the language models, including   1) \textbf{BERT~\cite{devlin2018bert} and BioBERT~\cite{lee2020biobert}}, 2) \textbf{GatorTron~\cite{yang2022gatortron}},  we also consider 3) \textbf{LLAMA family} as the baselines: MedLLaMA 13B~\cite{wu2023pmc}, LLaMA2-13b~\cite{touvron2023llama}.   5) \textbf{GPT-4}. \textbf{6) RA-\textsc{KNN}-$n$}: As shown in Figure~\ref{con:KNN_retrival},   LLaMA2 13B serves as the base model in UMLS and ADInt.


 \subsection{Evaluation Metrics}
In the Triple Extraction task, same as \cite{tang2022unirel,zeng2019learning}, triple is regarded as correct when its relation type, the
head entity and the tail entity are all correct. For example, in the sentence: \textit{Infusion of prostacyclin (PGI2) reportedly attenuates renal ischemic injury in the dog and the rat.},  triple \textit{<Infusion, treats, rat>} is regarded as correct while \textit{ <injury, treats, rat>} is not. 
Following the evaluation method of the previous work~\cite{tang2022unirel,shang2022onerel,lu2022unified,gao2023easy},  we evaluated all the models and reported the evaluation metric, including Micro Precision, Recall, and F1-score. For the relation extraction, text classification, and link prediction task, we follow the same evaluation metrics as triple extraction.

%% file: 7_conclution.tex
\section{Conclusion}
In this paper, we introduce a novel biomedical RAG framework called \textsc{BiomedRAG}. Unlike the traditional retrieval-argument language model, our framework retrieval the knowledge from the diverse chunk database and adapts the tailored chunk scorer to the LLM.
Experimental results show that our framework achieves consistent improvements in 4 biomedical NLP tasks over 8 datasets.





\section{Code and data availability}

The complete code and data will be  available in the repository:

--\url{https://github.com/ToneLi/PETAILOR-for-bio-triple-extraction}

\section{Acknowledgments}
This work was supported by the National Institutes of Health’s National Center for Complementary and Integrative Health grant number R01AT009457, National Institute on Aging grant number R01AG078154 and National Cancer Institute grant number R01CA287413. The content is solely the responsibility of the authors and does not represent the official views of the National Institutes of Health. 
Thanks to Huixue Zhou for suggesting revisions to the method section. Thanks to Chad Dupuis for solving the issues with our GPU server.


\section{Competing Interests}
The authors declare no competing financial or non-financial interests.